\documentclass{article}

\usepackage{microtype}
\usepackage{graphicx}
\usepackage{subfigure}
\usepackage{booktabs}

\usepackage{hyperref}

\usepackage[accepted]{icml2021}

\usepackage[utf8]{inputenc}
\icmltitlerunning{Convergence of Learning Algorithms Under Optimal Evaluative Feedback}

\newcommand{\prob}{\mathbb{P}}
\newcommand{\expect}{\mathbb{E}}
\newcommand{\given}{ \, | \, }
\newcommand{\namecite}[1]{\citeauthor{#1}~(\citeyear{#1})}

\def\fa{\forall \>}

\def\endproof{\newline \strut\hfill $\Box$}

\usepackage{amsmath}
\usepackage{amssymb}

\newcounter{thmno}
\newcommand{\thmNum}[1]{\refstepcounter{thmno}\label{#1}}
\newcommand{\set}[1]{\{#1\}}  
\newcommand{\N}{\mathbb{N}}  

\usepackage{tikz}
\usetikzlibrary{positioning}
\definecolor{processblue}{cmyk}{0.96,0,0,0}

\begin{document}
\twocolumn[
\icmltitle{Convergence of a Human-in-the-Loop Policy-Gradient Algorithm With Eligibility Trace Under Reward, Policy, and Advantage Feedback}

\icmlsetsymbol{equal}{*}

\begin{icmlauthorlist}
\icmlauthor{Ishaan Shah}{equal,brown}
\icmlauthor{David Halpern}{equal,brown}
\icmlauthor{Kavosh Asadi}{brown}
\icmlauthor{Michael L.\ Littman}{brown}

\end{icmlauthorlist}

\icmlaffiliation{brown}{Department of Computer Science, Brown University, Providence, Rhode Island, USA}

\icmlkeywords{Reinforcment Learning, ICML, Machine Learning, Convergence, COACH}

\icmlcorrespondingauthor{Ishaan Shah}{ishaan\_shah@alumni.brown.edu}
\icmlcorrespondingauthor{David Halpern}{david\_halpern@alumni.brown.edu}

\vskip 0.3in
]

\printAffiliationsAndNotice{\icmlEqualContribution}

\begin{abstract}
\textit{Abstract}---Fluid human--agent communication is essential for the future of human-in-the-loop reinforcement learning. An agent must respond appropriately to feedback from its human trainer even before they have significant experience working together. Therefore, it is important that learning agents respond well to various feedback schemes human trainers are likely to provide.  This work analyzes the COnvergent Actor--Critic by Humans (COACH) algorithm under three different types of feedback---policy feedback, reward feedback, and advantage feedback.  For these three feedback types, we find that COACH can behave sub-optimally. 
We propose a variant of COACH, episodic COACH (E-COACH), which we prove converges for all three types.
We compare our COACH variant with two other reinforcement-learning algorithms: {Q}-learning and TAMER.
\end{abstract}
\section{Introduction}
\label{introduction}
We study the algorithm COACH~\cite{coach}, designed to learn from evaluative feedback.
We would like for the algorithm to find an optimal policy under different feedback schemes, since a human trainer is apt to select from several possible approaches and we do not know which will be chosen \emph{a priori}. 

We present a proof of convergence for three natural feedback schemes. 
1) Feedback can take the form of an economic incentive in which the learner gets an immediate reward for moving into a state based on the state's desirability---one-step {\bf reward}. 2) Feedback can be a binary signal that tells the learner whether the action it took was correct (1) or not (0) with respect to the trainer's intended {\bf policy}. And, 3) feedback can reveal how good an action was relative to the agent's recent behavior---the action's {\bf advantage}. It is desirable for a learning algorithm to perform appropriately in all three of these settings.


E-COACH (Algorithm~\ref{coachAlgorithm}) is such a learning algorithm. It takes input policy $\pi_\theta$, discount factor $\gamma$, and a learning rate $\alpha$.

\begin{algorithm}[h]
\caption{E-COACH $\langle \pi_\theta, \gamma, \alpha \rangle$}
\label{coachAlgorithm}
\begin{algorithmic}

\STATE $\theta_0 \gets 0$
\FOR{$\textnormal{episode} = 0, 1, 2, \ldots$}
    \STATE $e_0 \gets 0$
    \FOR{$t = 0, 1, 2, \ldots$}
        \STATE $a _t \sim \pi_\theta (s_t,\cdot)$
        \STATE observe state $s_{t + 1}$ and human feedback $f_{t + 1}$
        \STATE $e_{t + 1} \gets e_{t} + \frac{1}{\pi_{\theta}(s_t, a_t)} \nabla_{\theta} \pi_{\theta}(s_t, a_t)$
        \STATE $\theta_{t + 1} \gets \theta_t + \alpha\gamma^t e_{t + 1} f_{t + 1}$
\ENDFOR
\ENDFOR
\end{algorithmic}
\end{algorithm}
E-COACH (Episodic COACH) is a close variant of the original COACH with a few differences. 1) It keeps track explicitly of the number of steps $t$ elapsed in the current episode. 2) E-COACH's most notable difference from COACH is E-COACH's use of a $\gamma^t$ decay factor. This element emphasizes information from temporally closer decisions over more distant ones. 3) In addition, E-COACH does not use a $\lambda$ parameter to decay the eligibility trace $e_t$. This makes E-COACH's treatment of eligibility traces like setting $\lambda=1$ in the original COACH algorithm.

We propose E-COACH instead of COACH because COACH does not take advantage of the discount factor, $\gamma^t$.  This causes it to incorrectly estimate the expected reward, causing it to perform poorly on the given environment.
We provide an example of such a scenario in \ref{originalCOACHOneStepRewardSection}. In contrast to COACH, we show that E-COACH can find converge under all three feedback schemes described above.
\section{Background}
\label{background}
A Markov Decision Process (MDP) is a five-tuple: $\langle S, A, T, R, \gamma \rangle$. Here, $S$ is a set of reachable states, $A$ is the set of actions an agent might use, $T(s' \given s, a)$ is a probability that the agent would move to state $s'$ from the given state $s$ having taken action $a$, $R(s, a)$ is the reward obtained for taking action $a$ from state $s$, and $\gamma \in [0, 1)$ is a discount factor indicating the importance of immediate rewards as opposed to rewards received in distant future.

A stochastic policy $\pi_\theta : S \times A \rightarrow [0, 1]$, where $\sum_{a \in A} \pi_{\theta}(s, a) = 1, \fa s \in S$, defines an agent's behavior via $\pi_\theta(s, a) = \prob\set{a_t = a \given s_t = s, \theta}, \fa s \in S, a \in A$.  Note that $\theta$ is a vector parameter of the policy, and we assume that $\pi$ is differentiable with respect to this parameter. 
For brevity, we will denote $\pi_\theta(s, a)$ as $\pi(s, a)$ when the parameter vector is clear from context.

The value functions $Q^\pi$ and $V^\pi$ measure the performance of policy $\pi$:
\begin{align*}
Q^\pi(s, a) = \expect\big[ \sum_{k = 1,2,\ldots} \gamma^{k - 1}r_{t + k} \given s_t = s, a_t = a, \pi \big]    
\end{align*}
and
\begin{align*}
    V^\pi(s) =  \expect_{a \sim \pi(s,\cdot)}[Q^{\pi}(s,a)\given s,\pi].
\end{align*}
When an agent's policy $\pi^* = \text{argmax}_{\pi} V^{\pi}, \fa s \in S$, then we call that policy \emph{optimal}.  We will denote optimal policies as $\pi^*$ and use the shorthand $V^{*}(s) = V^{\pi^*}(s)$. Also, for sake of brevity, we will write $\expect[\, \cdot \given \pi]$ simply
as $\expect[\cdot]$ from now on. Note that all expectations we consider are conditioned on the policy. If not specified otherwise
, $\expect[\cdot]$ is an expectation over $s_1, a_1 ,s_2 ,a_2 \ldots$ where $s_{t+1} \sim \text{T}(\cdot \given s_t,a_t)$ and $a_{t+1} \sim \pi(s_{t+1},\cdot)$.

\section{E-COACH Under Reward Feedback \label{oneStepRewardSection}}

A simple form of feedback a trainer may choose to give a learner is the one-step reward obtained from the MDP for the action the agent just took. Such \emph{reward feedback} is convenient 
since it is myopic and does not require the trainer to consider future rewards. It assumes a direct analogy between the rewards that define the task and the feedback provided by the trainer---it is the simplest extension of standard reinforcement learning to the interactive setting.  For the following theoretical results to hold, we assume the human-trainer gives consistent reward, as per the definition of our feedback, $f$.  $f$ represents our feedback, which we will redefine in Sections \ref{oneStepRewardSection}, \ref{indicatorSection}, and \ref{advantageSection}.

We look at an MDP $M = \langle S,A,R,T,\gamma \rangle$. Under reward feedback, when an agent takes an action $a$ in state $s$, the trainer gives feedback
\begin{align*}
    f(s, a) = R(s, a).
\end{align*}

\thmNum{oneStepThmNum} 
\textbf{Theorem \ref{oneStepThmNum}:} E-COACH converges under reward feedback $f(s, a) = R(s, a), \fa s \times a \in S \times A$.

\textbf{Proof:} Consider the sequence of updates on $e_t$ and $\theta_{t}$ at each time step $t$:
\begin{align*}
e_{t+1} &\gets e_{t} +
\frac{1}{\pi_{\theta}(s_t, a_t)} \nabla_{\theta} \pi_{\theta}(s_t, a_t) \\
\theta_{t + 1} &\gets \theta_t + \gamma^t e_{t+1} r_{t + 1}
\end{align*}
where, for brevity, we define $r_{t + 1} = R(s_{t},a_{t})$.

To better understand what the updates mean, consider some terminal time $L$. The value $L$ may refer to the time at which an agent reaches the goal or a pre-decided time at which the trainer stops the agent. We use $L$ only for the purpose of elucidation and the analysis below also extends to the infinite horizon case when $L$ is unbounded. We ignore the $\alpha$ in the $\theta$ update above for sake of clarity. By linearity of the updates, it is trivial to incorporate $\alpha$ into the calculations below.
\begin{align*}
\theta_{L+1} &= \Sigma_{\tau = 0}^{L} \gamma^{\tau}e_{\tau + 1} r_{\tau +1}\\
&= \Sigma_{\tau = 0}^{L} \gamma^{\tau}r_{\tau +1}(\Sigma_{t = 0}^{\tau} \frac{\nabla_{\theta}\pi_{\theta}(s_t,a_t)}{\pi_{\theta}(s_t, a_t)})\\
&= \Sigma_{\tau = 0}^{L} \Sigma_{t = 0}^{\tau}  \gamma^{\tau}r_{\tau +1} \frac{\nabla_{\theta}\pi_{\theta}(s_t,a_t)}{\pi_{\theta}(s_t, a_t)}
\end{align*}
Rearranging the order of summation
\begin{align*}
\theta_{L + 1} &=  \Sigma_{t = 0}^{L} \Sigma_{\tau = t}^{L} \frac{\nabla_{\theta}\pi_{\theta}(s_t,a_t)}{\pi_{\theta}(s_t, a_t)} \gamma^{\tau} r_{\tau +1} \\
&=  \Sigma_{t = 0}^{L}\gamma^t \frac{\nabla_{\theta}\pi_{\theta}(s_t,a_t)}{\pi_{\theta}(s_t, a_t)} (\Sigma_{\tau = 0}^{L-t}  \gamma^{\tau} r_{\tau + t + 1})
\end{align*}
Taking expectation
\begin{align*}
 \expect[\theta_{L+1}] = \Sigma_{t = 0}^{L} \gamma^{t}\expect[\frac{\nabla_{\theta}\pi_{\theta}(s_t,a_t)}{\pi_{\theta}(s_t, a_t)} Q^{\pi}(s_{t},a_{t})]
\end{align*}

\subsection{E-COACH Objective Function}
\label{obj-func}

In this section we show that the gradient of the objective function $\Sigma_{t = 0}^{\infty}\gamma^{t}\expect[\frac{\nabla_{\theta}\pi_{\theta}(s_t,a_t)}{\pi_{\theta}(s_t, a_t)} Q^{\pi}(s_{t},a_{t})]
$ is what REINFORCE performs gradient ascent on. Note that this quantity is what E-COACH is estimating (via $\theta_L$ parameter) and performing gradient ascent on.

Consider the REINFORCE algorithm (Algorithm \ref{reinforceAlgorithm}). Here, $G_{t}$ is a Monte Carlo estimate of $Q^{\pi}(s_t,a_t)$. Hence, at any terminal time $L$, it is clear that the expected value of $\theta_{L}$ obtained by REINFORCE is equal to that of E-COACH. 
\begin{algorithm}[h]
\caption{ REINFORCE$<\pi_\theta, \gamma, \alpha>$}
\label{reinforceAlgorithm}
\begin{algorithmic}
\STATE Generate an episode $s_0, a_0, r_1, \ldots s_{L}, a_{L}, r_{L+1}$

\FOR{$t = 0, 1, 2, \ldots$}
    \STATE $G_{t}$ = return from step t
    \STATE $\theta_{t + 1} \gets \theta_t + \gamma^t G_{t}
    \nabla_\theta \log (\pi_\theta(s_t, a_t))$
\ENDFOR
\end{algorithmic}
\label{reinforce}
\end{algorithm}

Consider the unnormalised state visitation distribution, such that $ \forall s \in S$, $d^\pi(s) = \prob_0^\pi(s) + \gamma \prob_1^\pi(s) + \ldots + \gamma^{i} \prob_{i}^\pi(s) + \ldots$
where $\prob_t^\pi(s)$ denotes the probability of arriving in state $s$ at time $t$ following policy $\pi$. 
The objective to maximize, as described in the Policy Gradient Theorem~\cite{sutton}, is 
\begin{align*}
\rho^{\pi} = \Sigma_{s} d^{\pi}(s) \Sigma_{a} \pi(s,a) R(s,a),
\end{align*}
and its gradient is
\begin{align*}
 \nabla_{\theta}\rho^{\pi} = \Sigma_{s} d^{\pi}(s) \Sigma_{a} \nabla_{\theta}\pi(s, a) Q^{\pi}(s,a).
\end{align*}
The gradient can be rewritten as $\nabla_{\theta} \rho^{\pi} = \expect_{s \sim d^{\pi},a \sim \pi(s,\cdot)}[\frac{\nabla_{\theta}\pi_{\theta}(s_t,a_t)}{\pi_{\theta}(s_t, a_t)}  Q^{\pi}(s,a)]$. 

Expanding with respect to $d^{\pi}(s)$ yields 
\begin{align*}
  \expect_{s \sim d^{\pi},a \sim \pi(s,\cdot)}[&\frac{\nabla_{\theta}\pi_{\theta}(s_t,a_t)}{\pi_{\theta}(s_t, a_t)}  Q^{\pi}(s,a)] \\ 
  = \Sigma_{t = 0}^{\infty}\gamma^{t}\expect_{s_t \sim \prob^{\pi}_t,a_t \sim \pi_{\theta}(s_t,\cdot)}[&\frac{\nabla_{\theta}\pi_{\theta}(s_t,a_t)}{\pi_{\theta}(s_t, a_t)} Q^{\pi}(s_{t},a_{t})]
\end{align*}

which is exactly what E-COACH and REINFORCE are estimating. 
Since E-COACH is performing gradient ascent on the policy gradient objective, we can use results from \cite{policyGradientConvergence} to say that E-COACH converges to local optima or saddle points.  Although, recent work has shown that policy gradient methods can escape saddle points under mild assumptions on the rewards and minor modifications to existing algorithms \cite{escapeSaddlePoints1} \cite{escapeSaddlePoints2}. \endproof

Something to note is that behavioral evidence indicates that one-step reward is not a typical choice of human trainers~\cite{markHoPaper}.

\section{E-COACH Under Policy Feedback}

To argue that E-COACH converges under policy feedback, we first consider a more general form of feedback and then show policy feedback is a special case.

\subsection{E-COACH with a More General Type of Feedback}
\label{generalization}

Let us start by considering two similar MDPs $M_1 = \langle S, A, R, T, \gamma\rangle$ and $M_2 = \langle S, A, f, T, \gamma\rangle$.  Note the differing reward functions $R$ and $f$ in the two MDPs.

We will denote the value functions for $M_1$ and $M_2$ as $V_1$ and $V_2$, respectively.  We will say that the starting state for both of our MDPs is $s_0$. Define $V_1^{\min} = \min_{\pi \in \Pi} V_1^\pi (s_0)$, $V_1^{*} = \max_{\pi \in \Pi} V_1^\pi(s_0)$.

The following theorem will have the following assumption:

\begin{enumerate}
    \item E-COACH (see algorithm \ref{coachAlgorithm}) will give us a policy $\pi_2(s, a)$ such that $\expect_{s \sim d^{\pi_2^*}} \big[\sum_a\big|\pi_2^*(s, a) - \pi_2(s, a)\big|\big] \leq \delta$ for some optimal policy $\pi_2^*$ on the domain $M_2$.  The proof in section \ref{oneStepRewardSection} strengthens this assumption by showing that E-COACH optimizes the policy gradient objective.  Note $\pi_2^*$ may not be the only optimal policy; instead, it is a single optimal policy. \label{kavoshAssumption}
    \item We also assume that $\gamma \neq 1$ for the case where the MDP has an infinite horizon, which will we will justify later on. \label{lambdaNotOneAssumption}
\end{enumerate}
Theorem \ref{genThmNum} requires the condition that all optimal policies for $M_2$ are also optimal for $M_1$.  We will later show that this condition holds true for the case of policy feedback in theorem \ref{indicatorThmNum}, allowing us to leverage these results.

\thmNum{genThmNum}
\textbf{Theorem~\ref{genThmNum}:} If all optimal policies for $M_2$ are also optimal for $M_1$ (optimal policies of $M_2$ are a subset of those for $M_1$), then running E-COACH on $M_2$ will result in a policy that is close to an optimal policy on $M_2$, which will also be close to an optimal policy for $M_1$.  Let's define $W = \max(|V_1^*|, |V_1^{\min}|)$.  Then we find that,
\begin{align*}
    0 \leq V_1^* - V_1^{\pi_2} \leq W \delta
\end{align*}
\textbf{Proof:} 
We have to show that running E-COACH in $M_2$ will yield a policy that is not too far off from an optimal policy for $M_1$.  We would like to run E-COACH on $M_2$, using the alternate form of feedback as the reward function, and for any good policy (as per assumption \ref{kavoshAssumption}) we get from E-COACH on $M_2$, we would like for that policy to also be good on $M_1$, the original MDP we are trying to solve.  

The lower-bound in the theorem statement is immediate.

For the upper-bound, let's let $\pi^{(n)}$ denote a policy that follows/simulates $\pi_2^*$ for the first $n - 1$ time-steps and $\pi_2$ for the rest.  Hence, on the $n^\textnormal{th}$ time-step, $\pi^{(n)}$ will follow/simulate $\pi_2$ and not $\pi_2^*$.  Let $V^{(n)}$ denote the value of policy $\pi^{(n)}$.  Therefore, we can say that $V_1^{\pi_2} = V^{(0)}$ and $V_1^* = V^{(\infty)}$. Remember that $\pi_{2}^{*}$ is optimal on $M_1$ \emph{and} $M_2$ by the condition above, and thus has value $V^{*}$.

We'll start by considering $V^{(t)} - V^{(t - 1)}$.  Both $\pi^{(t)}$ and $\pi^{(t - 1)}$ accumulate the same expected reward for the first $t - 2$ steps and so these rewards cancel out.  Note that the $\prob$ we use below is the same as that defined in section \ref{obj-func}.  We find the following:
\begin{align*}
    V^{(t)} - &V^{(t - 1)} = \gamma^{t - 1} \sum_s \prob_{t - 1}^{\pi_2^*}(s) \sum_a \pi_2^*(s,a) Q^{\pi_2}(s, a) \\
    - &\gamma^{t - 1} \sum_s \prob_{t - 1}^{\pi_2^*}(s) \sum_a \pi_2(s, a) Q^{\pi_2}(s, a) \\
    = &\gamma^{t - 1} \sum_s \prob_{t - 1}^{\pi_2^*}(s) \sum_a (\pi_2^*(s, a) - \pi_2(s, a)) Q^{\pi_2}(s, a) 
    \\
    \leq &\gamma^{t - 1} \sum_s \prob_{t - 1}^{\pi_2^*}(s) \sum_a | \pi_2^*(s, a) - \pi_2(s, a) | W
\end{align*}
Now we'll use the above fact when considering $V_1^* - V_1^{\pi_2}$.
\begin{align*}
    V_1^* - V_1^{\pi_2} =& (V^1 - V^0) + (V^2 - V^1) + (V^3 - V^2) + \cdots \\
    \leq & \sum_{i}^\infty \gamma^i \sum_s \prob_i^{\pi^*_2}(s) \sum_a |\pi_2^*(s, a) - \pi_2(s, a)| W \\
    = &  \sum_s \sum_{i}^\infty \gamma^i \prob_i^{\pi^*_2}(s) \sum_a |\pi_2^*(s, a) - \pi_2(s, a)| W \\
    = & \sum_s d^{\pi_2^*}(s) \sum_a |\pi_2^*(s, a) - \pi_2(s, a)| W \\
    = & W \expect_{s \sim d^{\pi_2^*}} \big[\sum_a|\pi_2^*(s, a) - \pi_2(s, a)|\big] \\
    \leq & W \delta
\end{align*}
As $\delta \rightarrow 0$, we have that $V_1^* - V_1^{\pi_2} \rightarrow 0$.
\endproof

Note that our theorem says something different than the Simulation Lemma \cite{sim-lemma} as we make no assumptions about how close the reward functions of $M_1$ and $M_2$ are. Instead our theorem requires optimal policies in $M_2$ be optimal in $M_1$ and bounds the return of a policy learnt by E-COACH in $M_2$.  

\subsection{E-COACH Under Policy Feedback}
\label{indicatorSection}


Let $M_1 = \langle S, A, R, T, \gamma \rangle$ be an MDP without any specific reward function.  Under \emph{policy feedback}, a trainer has a target stationary deterministic policy $\pi^*_1$ in mind and delivers feedback based on whether the trainer's decision agrees with $\pi^*_1$.
When an agent takes an action $a$ in state $s$, the trainer will give feedback
\begin{align*}
    f(s, a) = I(s,a),
\end{align*}
with $I(s,a)$ defined as, 
\begin{align*}
    I(s, a) = 
    \begin{cases}
        1, & \textnormal{if } \pi^*_1(s) = a,\\
        0, & \textnormal{otherwise.}
    \end{cases}
\end{align*}

\thmNum{indicatorThmNum}
\textbf{Theorem \ref{indicatorThmNum}:} E-COACH converges under feedback $f(s, a) = I(s,a),\fa s\times a \in S \times A$.

\textbf{Proof:}
Consider the case of replacing the reward function $R(s,a)$ with $I(s,a)$ in MDP $M_{1}$, constructing a new MDP $M_2 = \langle S, A, f, T, \gamma \rangle$.  We would like to show that, in this setting, the E-COACH algorithm
converges to the optimal solution.  
$M_1$ and $M_2$ satisfy the prerequisites for theorem \ref{genThmNum}.

Consider the optimal policy for $M_2$. The best policy will select the best action in every state.  We have that $V^*_2(s_0) = \sum_{i = 0}^\infty 1 \cdot \gamma^i$.  The optimal policy for $M_2$ will achieve this value 
function because, if not, then we have a policy such that $V_2'(s_0) = \sum_{i = 0}^\infty t(i) \cdot 1 \cdot \gamma^i$, 
where $t(i) \in \set{0, 1} \fa i$ and $t(j) = 0$ for some $j$.  
Take the smallest value $k \in \N$ such that $t(k) = 0$, then $V^*_2 (s_0) - V_2'(s_0) \geq \gamma^k$ 
so then the policy achieving $V_2'$ is sub-optimal.  We can conclude that $V^*_2(s_0)$ is the value function for the optimal policy.  Therefore, the policy that always chooses the action that gives a value of one is optimal.  Also, note that always choosing the action that results in a feedback of one corresponds exactly to the decision of $\pi_1^*$ by construction of $f(s, a)$.  So, we obtain that $\pi_1^*(s, a) = \pi_2^*(s, a), \fa (s, a) \in S \times A$.  In other words, an optimal policy in the new domain is equivalent to the target policy from the original one.

We can leverage Theorem~\ref{genThmNum} to show that the algorithm converges under policy feedback.
\endproof

\section{E-COACH Under Advantage Feedback}
\label{advantageSection}

COACH~\cite{coach} was originally motivated by the observation that human feedback is observed to be policy dependent---if a decision \emph{improves} over the agent's recent decisions, trainers provide positive feedback. If it is worse, the trainer is more likely to provide negative feedback. As such, feedback is well modeled by the advantage function of the agent's current policy.

In \emph{advantage feedback}, when an agent takes action $a$ in state $s$, the trainer will give feedback
\begin{align*}
    f(s, a) = A^{\pi}(s,a),
\end{align*}
with $A^{\pi}(s,a)$ defined as, 
\begin{align*}
    A^{\pi}(s, a) = Q^{\pi}(s,a) - V^{\pi}(s).
\end{align*}
\thmNum{advThmNum}
\textbf{Theorem \ref{advThmNum}:} E-COACH converges under feedback $f(s, a) = A^{\pi}(s,a),\fa s\times a \in S \times A$.

\textbf{Proof:}
Since $V^{\pi}(s) =  \expect_{a \sim \pi(\cdot |s)}[Q^{\pi}(s,a) \,|s]$, we have that 

\begin{align*}
    \expect_{a \sim \pi(s, \cdot)}[
&A^{\pi}(s,a) \given s] \\ 
=& \expect_{a \sim \pi(s, \cdot)}[
Q^{\pi}(s,a) \given s]  - V^{\pi}(s) \\
=& 0
\end{align*}
By the equation above we can say the following
\begin{equation}
    \expect_{s_{t+1},a_{t+1},s_{t+2} \ldots}[A^{\pi}(s_{t+\tau},a_{t+\tau})|s_t,a_t] = 0 \, \, \forall \tau > 0 \label{expectOfAdvEqualsZero}
\end{equation}
We also know that
\begin{equation}
    \expect_{s_{t+1},a_{t+1},s_{t+2} \ldots}[A^{\pi}(s_{t},a_{t})|s_t,a_t] = A^{\pi}(s_{t},a_{t}) \label{expectOfAdvEqualsAdv}
\end{equation}
We will use equations \ref{expectOfAdvEqualsZero} and \ref{expectOfAdvEqualsAdv} later on in this proof.

Using the same approach as in theorem \ref{oneStepThmNum}, we look at the sequence of updates made to the policy parameter $\theta$ until some terminal time $L$.

\begin{align*}
\theta_{L+1} &= \Sigma_{\tau = 0}^{L} \gamma^{\tau}e_{\tau + 1}A^{\pi}(s_{\tau},a_{\tau})\\
&= \Sigma_{\tau = 0}^{L} \gamma^{\tau}A^{\pi}(s_{\tau},a_{\tau})(\Sigma_{t = 0}^{\tau} \frac{\nabla_{\theta}\pi_{\theta}(s_t,a_t)}{\pi_{\theta}(s_t, a_t)})\\
&= \Sigma_{\tau = 0}^{L} \Sigma_{t = 0}^{\tau}  \gamma^{\tau}A^{\pi}(s_{\tau},a_{\tau}) \frac{\nabla_{\theta}\pi_{\theta}(s_t,a_t)}{\pi_{\theta}(s_t, a_t)}
\end{align*}
Rearranging the order of summation
\begin{align*}
\theta_{L + 1} &=  \Sigma_{t = 0}^{L} \Sigma_{\tau = t}^{L} \frac{\nabla_{\theta}\pi_{\theta}(s_t,a_t)}{\pi_{\theta}(s_t, a_t)} \gamma^{\tau}A^{\pi}(s_{\tau},a_{\tau}) \\
&=  \Sigma_{t = 0}^{L}\gamma^t \frac{\nabla_{\theta}\pi_{\theta}(s_t,a_t)}{\pi_{\theta}(s_t, a_t)} (\Sigma_{\tau = 0}^{L-t}  \gamma^{\tau} A^{\pi}(s_{\tau + t},a_{\tau + t}))
\end{align*}
Therefore, taking the expectation
\begin{align*}
  \expect[\theta_{L+1}] &= \Sigma_{t = 0}^{L} \gamma^{t}\expect[\frac{\nabla_{\theta}\pi_{\theta}(s_t,a_t)}{\pi_{\theta}(s_t, a_t)} (\Sigma_{\tau = 0}^{L-t}  \gamma^{\tau} A^{\pi}(s_{\tau + t},a_{\tau + t}))]\\
 &= \, \Sigma_{t = 0}^{L} \gamma^{t} \expect_{s_t,a_t}[\frac{\nabla_{\theta}\pi_{\theta}(s_t,a_t)}{\pi_{\theta}(s_t, a_t)} \times \\
  &\Sigma_{\tau = 0}^{L-t}\gamma^{\tau}\expect_{s_{t+1},a_{t+1},s_{t+2} \ldots}[A^{\pi}(s_{\tau + t},a_{\tau + t})|s_t,a_t]]
\end{align*}
Using equations \ref{expectOfAdvEqualsZero} and \ref{expectOfAdvEqualsAdv}, we can say that
\begin{align*}
   \expect[\theta_{L + 1}] &=  \Sigma_{t = 0}^{L} \gamma^{t}\expect[\frac{\nabla_{\theta}\pi_{\theta}(s_t,a_t)}{\pi_{\theta}(s_t, a_t)} A^{\pi}(s_{t},a_{t})]
\end{align*}
Using the fact that $\expect[\frac{\nabla_{\theta}\pi_{\theta}(s_t,a_t)}{\pi_{\theta}(s_t, a_t)}V^{\pi}(s_t)] = 0$ \cite{baseline}
\begin{align*}
\expect[\theta_{L+1}] = \Sigma_{t = 0}^{L} \gamma^{t}\expect[\frac{\nabla_{\theta}\pi_{\theta}(s_t, a_t)}{\pi_{\theta}(s_t, a_t)} Q^{\pi}(s_t, a_t)] 
\end{align*}

Thus, using the argument in \ref{obj-func}, we can state that E-COACH converges under advantage feedback.  \endproof 

\begin{figure*}[h]
    \subfigure[]{\includegraphics[width=0.32\textwidth]{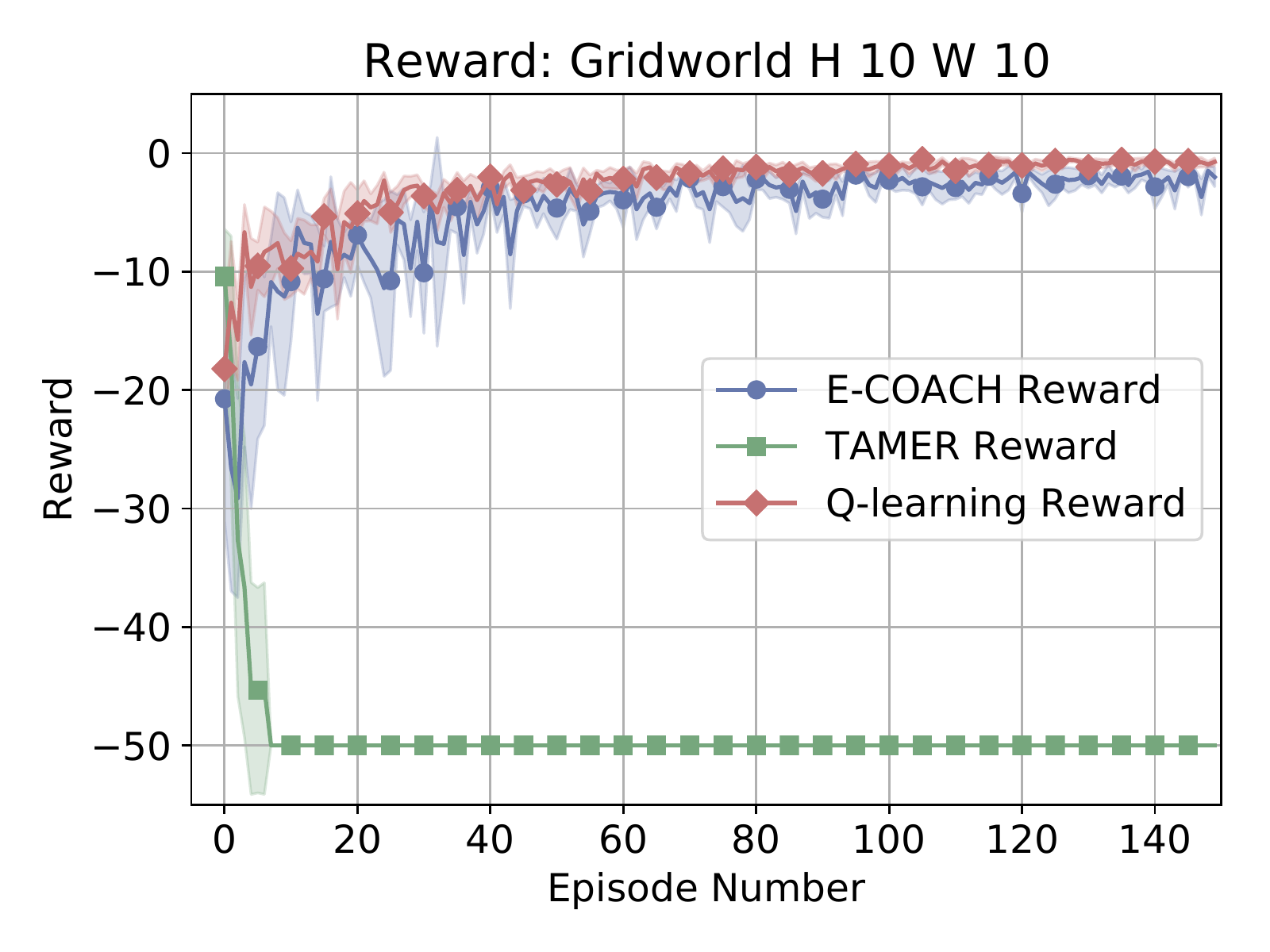}\label{graphReward}}
    \subfigure[]{\includegraphics[width=0.32\textwidth]{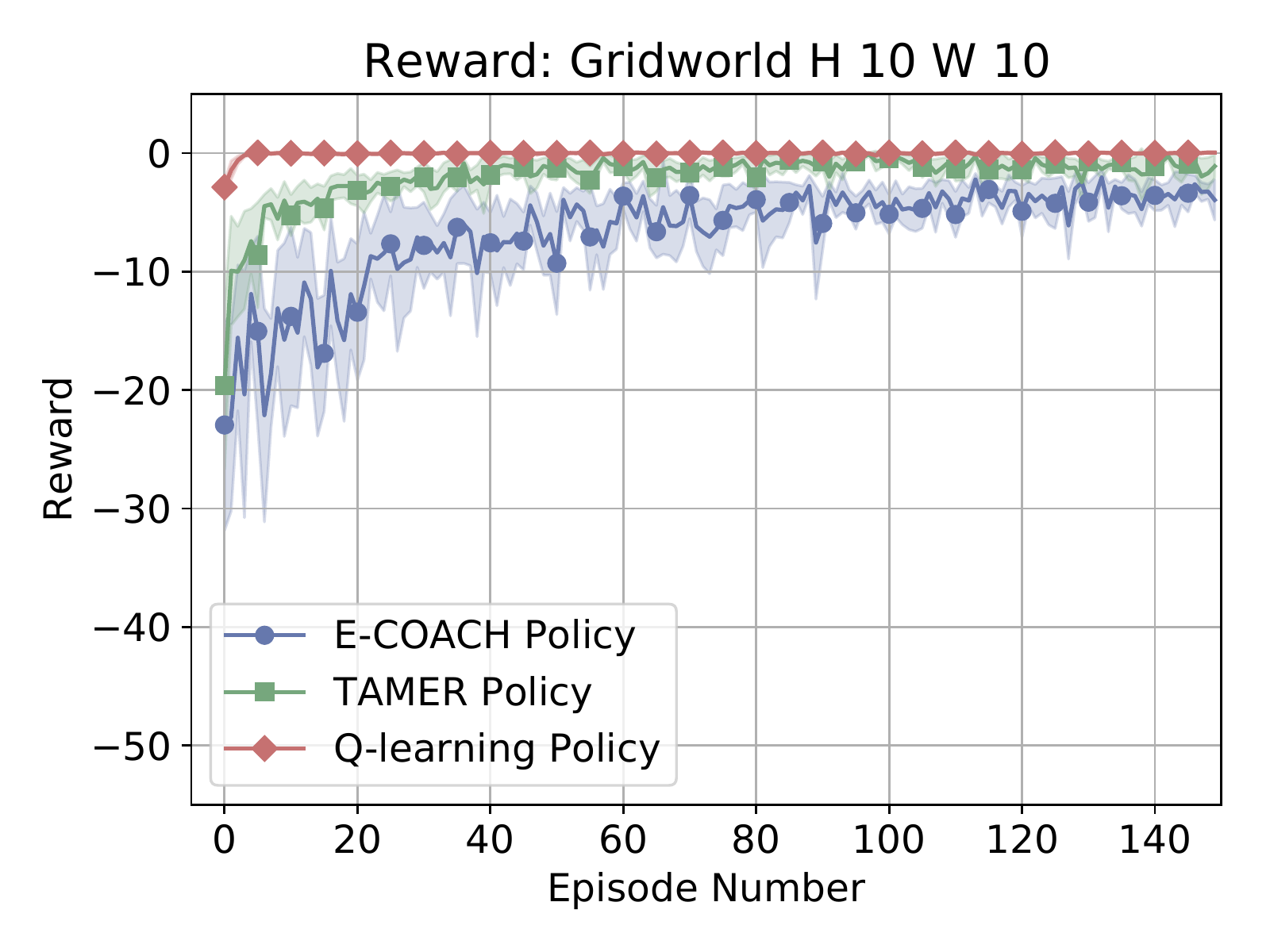}\label{graphPolicy}}
    \subfigure[]{\includegraphics[width=0.32\textwidth]{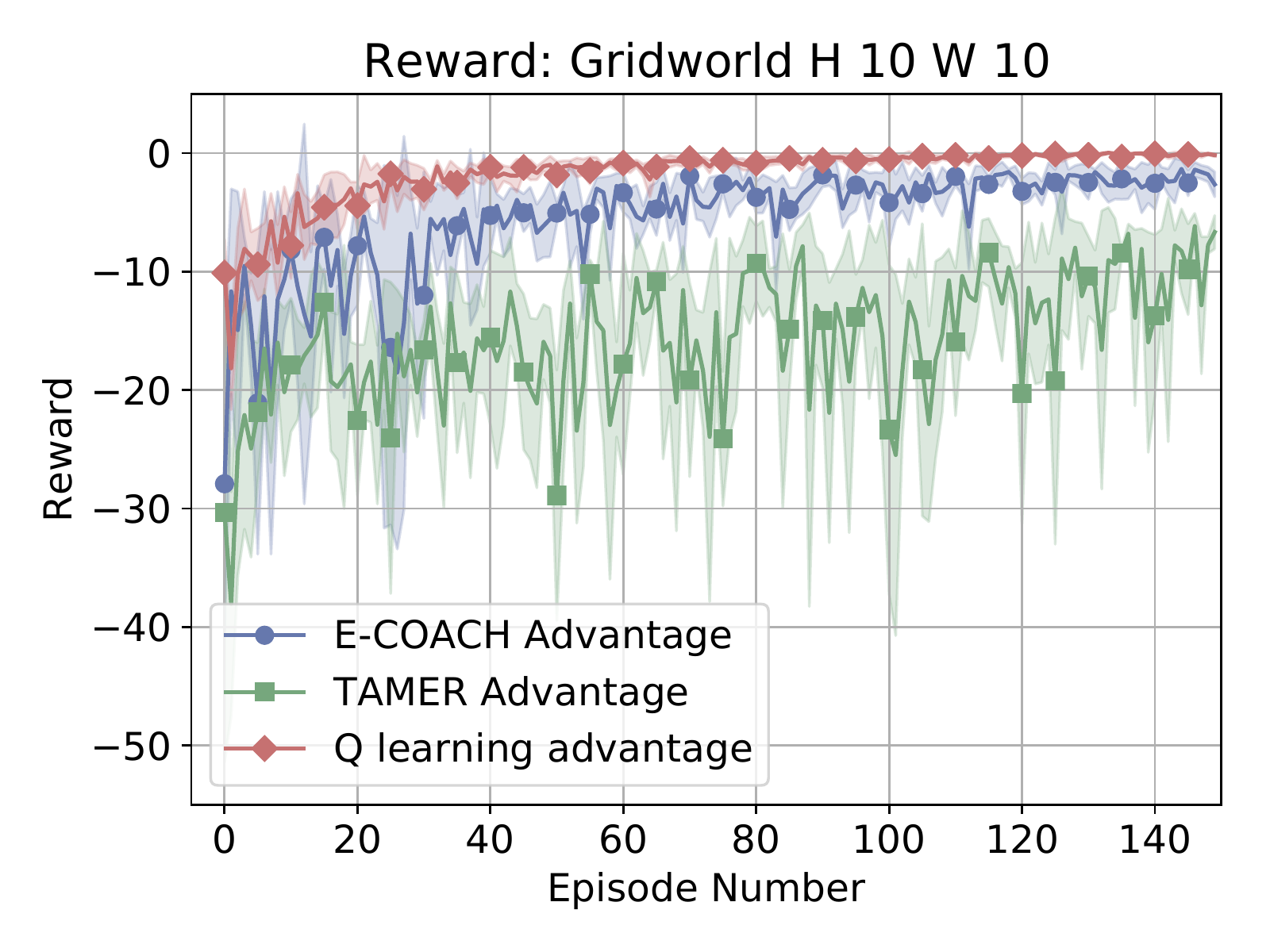}\label{graphAdvantage}}
    \caption{Performance of E-COACH, TAMER and Q-learning under each feedback scheme. The domain is a 10 by 10 GridWorld coded using simple\_rl \cite{abel2019spirl}.  Each agent was run for 150 episodes and was cut short after 1000 steps.  We ran 10 instances of each agent and plotted the reward averaged over these instances. E-COACH and Q-learning maximize rewards in all three settings, while TAMER falters under reward feedback and has difficulty with advantage.  Although, other experimental results show TAMER doing well with advantage feedback; see \citeauthor{macglashan2017interactive} (\citeyear{macglashan2017interactive}).  These experimental results are meant to support our proofs of convergence for E-COACH.  In addition, they support arguments made in Section \ref{comparisonSection}.}
    \label{graphFig}
\end{figure*}
\section{Original COACH}
This section assesses the convergence of the original COACH algorithm~\cite{coach} under the three different types of feedback defined in this paper.  Recall the main differences between E-COACH and COACH:
\begin{enumerate}
    \item COACH makes use of an eligibility decay factor $\lambda$. \label{lambdaDifference}
    \item COACH does not discount the feedback by $\gamma^t$ as part of the algorithm. \label{gammaDifference}
\end{enumerate}
Note that $\lambda$ and $\gamma$ are not replaceable as the $\lambda$ can only be used to discount stored gradients and thereby discount future rewards. On the other hand, $\gamma$ is used to both discount future rewards as well as estimate the unnormalised state visitation distribution $d^{\pi}(s)$ described in \ref{obj-func}.

As a result, the original COACH is incapable of estimating the state visitation distribution. Hence, the updates made at $t = 0$ and $t = 10$ would be weighted equally by COACH. 

This property goes against what policy-gradient algorithms would do. By not using $\gamma$, COACH is basically drawing from a state visitation distribution different from the state visitation distribution $d^{\pi}(s)$ that is part of the objective function $\rho$ that we described in \ref{obj-func}. As a result, the updates made by COACH are not estimating $\rho$. Although COACH may learn to do reasonably well, we cannot say that it will behave optimally.

\subsection{COACH Under One-Step Reward} \label{originalCOACHOneStepRewardSection}

The algorithm will converge, but the policy it converges to will be suboptimal for $\gamma \neq 1$ because COACH does not estimate the gradient of the policy gradient objective, $\nabla_{\theta}\rho$. Because COACH does not incorporate the discount factor, it behaves as if the domain has $\gamma = 1$, even if it isn't necessarily the case.  If the domain has a $\gamma \in [0, 1)$, then the policy will have a long-term view because it will ignore this discount factor. Ignoring discounts can lead to suboptimal behavior. 

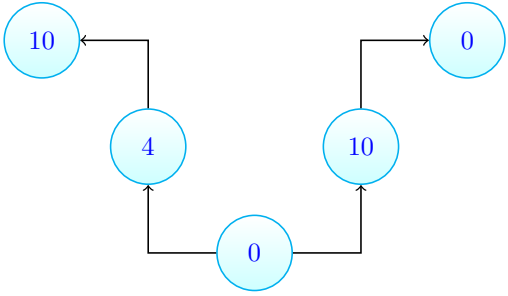
\begin{figure}[h]
\begin{center}
\begin{tikzpicture}[-latex, auto, node distance=2cm, on grid, semithick, state/.style ={ circle ,top color =white , bottom color = processblue!20 ,
draw,processblue , text=blue , minimum width =1 cm}]
\node[state] (A) {$0$};
\node[state] (B) [above left =of A] {$4$};
\node[state] (C) [above left =of B] {$10$};
\node[state] (D) [above right =of A] {$10$};
\node[state] (E) [above right =of D] {$0$};
\draw[->] (A) -| (B);
\draw[->] (A) -| (D);
\draw[->] (B) |- (C);
\draw[->] (D) |- (E);
\end{tikzpicture}
\end{center}
\caption{Example of the impact of discount factor on optimal policies.  The number in the circle represents reward.  The state in the middle is our starting state.  For $\gamma \approx 1$, it is clear that the optimal  policy is to go left to obtain a value of $\approx 14$.  For $\gamma \approx 0$, the optimal actions is to go right, instead, to obtain a value of $\approx 10$ instead of $\approx 4$. In general, the left action is preferred for $\gamma > 0.6$ and the right action is preferred for $\gamma < 0.6$. The choice of $\gamma$ impacts the optimal policy.}
\label{originalCOACHOneStepRewardCounterExampleFigure}
\end{figure}

Consider the five-state domain in Figure \ref{originalCOACHOneStepRewardCounterExampleFigure}. It shows how the optimal decision in a state can change with the discount factor. 

\subsection{Policy Feedback}

COACH will converge, but to a poor performing policy for the same reason given in Section \ref{originalCOACHOneStepRewardSection}.

\subsection{Advantage Feedback}


COACH should converge under this feedback type as per the argument given by \citeauthor{coach}.


\section{Comparison With Other Algorithms}
\label{comparisonSection}

We now know that E-COACH converges under several types of feedback. The three highlighted in this paper are Policy, Advantage, and Reward feedback.  In this section, we compare E-COACH to TAMER~\cite{tamer} and {Q}-learning under these three types of feedback.

\subsection{TAMER}

TAMER expects the human trainer to take each action's long-term implications into account when providing feedback.  TAMER learns the trainer's feedback function, then returns the policy that maximizes one-step feedback in each state.

The pseudocode as described in algorithm \ref{tamerRunAgent} is the TAMER algorithm.  See \cite{tamer} for more details.  The $t$ represents the time, the weights $\overrightarrow{w}$ is used for the reward model, and the feature vectors $\overrightarrow{f_{t - 2}}$ and $\overrightarrow{f_{t - 1}}$ are state feature vectors.  It takes input $\alpha$, a learning rate.

Note that there are several different versions of TAMER.  The one we are analyzing is the original by \citeauthor{tamer} (\citeyear{tamer}).

\begin{algorithm}[h]
\caption{TAMER $\langle \alpha \rangle$} \label{tamerRunAgent}
\begin{algorithmic}
\STATE $t \gets 0$
\STATE $\overrightarrow{w} \gets \overrightarrow{0}$
\STATE $\overrightarrow{f_{t - 2}} \gets \overrightarrow{0}$
\STATE $\overrightarrow{f_{t - 1}} \gets \overrightarrow{0}$
\STATE $a \gets ChooseAction(s_t, \overrightarrow{w})$
\STATE takeAction(a)
\WHILE {true}
    \STATE $t \gets t + 1$
    \IF {$t \geq 2$}
        \STATE $r_{t - 2} \gets getHumanFeedback()$
        \IF {$r_{t - 2} \neq 0$}
            \STATE $\overrightarrow{w} \gets UpdateRewModel(r_{t - 2}, \overrightarrow{f_{t - 2}}, \overrightarrow{f_{t - 2}}, \overrightarrow{w}, \alpha)$
        \ENDIF
    \ENDIF
    \STATE $a \gets ChooseAction(s_t, \overrightarrow{w})$
    \STATE $takeAction(a)$
    \STATE $s_t \gets getState()$
    \STATE $f_{t - 2} \gets f_{t - 1}$
    \STATE $f_{t - 1} \gets getFeatureVec(s_t)$
\ENDWHILE
\end{algorithmic}
\end{algorithm}

\begin{algorithm}[h]
\caption{UpdateRewModel $\langle r_{t - 2}, \overrightarrow{f_{t - 2}}, \overrightarrow{f_{t - 1}}, \overrightarrow{w}, \alpha \rangle$}
\begin{algorithmic}
\STATE Set $\alpha$ as a parameter.
\STATE $\overrightarrow{\Delta f_{t-1, t-2}} \gets \overrightarrow{f_{t - 1}} - \overrightarrow{f_{t - 2}}$
\STATE $projectedRew_{t - 2} \gets \sum_i (w_i \times \overrightarrow{\Delta f_{t-1, t-2}})$
\STATE $error \gets r_{t - 2} - projectedRew_{t - 2} $
\FOR{$i$ in $range(0, length(\overrightarrow{w}))$}
    \STATE $w_i \gets w_i + \alpha \times error \times \overrightarrow{\Delta f_{t-1, t-2}}$
\ENDFOR
\STATE \textbf{return } $\overrightarrow{w}$
\end{algorithmic}
\end{algorithm}

\begin{algorithm}
\caption{ChooseAction $\langle s_t, \overrightarrow{w} \rangle$}
\begin{algorithmic}
\STATE $\overrightarrow{f_t} \gets getFeatureVec(s_t)$
\FOR{each $a \in getAction(s_t)$}
    \STATE $s_{t + 1, a} \gets T(s_t, a)$
    \STATE $\overrightarrow{f_{t + 1, a}} \gets getFeatureVec(s_{t + 1}, a)$
    \STATE $\overrightarrow{\Delta f_{t + 1, t}} \gets \overrightarrow{f_{t + 1, a}} - \overrightarrow{f_t}$
    \STATE $projectedRew_a \gets \sum_i (w_i \times \delta f_{t + 1, t, i}$
\ENDFOR
\STATE \textbf{return } $argmax_a(projectedRew_a)$
\end{algorithmic}
\end{algorithm}

\subsubsection{Reward Feedback}

Because TAMER will maximize over the learned function, it will result in a bad policy for this form of feedback.  TAMER does not take future rewards into account and instead will greedily maximize for immediate reward.  TAMER assumes the trainer has taken future rewards into account already.  See figure \ref{graphReward}.

\subsubsection{Policy Feedback}

TAMER expects policy feedback and chooses correct actions assuming sufficient exploration.  See figure \ref{graphPolicy}.

\subsubsection{Advantage Feedback}

It is not known precisely how TAMER responds to advantage feedback. 
\namecite{tamer} claim that TAMER should work under moving feedback.  That is, TAMER should behave properly even when feedback changes over time because the algorithm expects the human trainer to be inconsistent and continues to update its choices even in the face of changes.  The advantage function assigns different values to actions as the policy is updated, so at different times it gets different values.  Assuming TAMER is able to learn this moving function, then a greedy one-step policy should be optimal because the maximal value of the advantage function is always the optimal action for the given state.  See figure \ref{graphAdvantage}.

\subsection{{Q}-learning}

{Q}-learning \cite{qlearning} is an algorithm that expects feedback in the form of immediate reward and calculates long-term value from these signals. Specifically, $Q_k(s,a)$ is its estimate of long-term value and, when it is informed of a transition from $s_t$ to $s_{t+1}$ via action $a_t$ and feedback $f_t$, it makes the update:
\begin{align*}
    Q_{k + 1}(s_t, a_t) \gets &(1 - \alpha) Q_k(s_t, a_t) \\
    &+ \alpha (f_t + \gamma V(s_{t + 1})).
\end{align*}

\subsubsection{Reward Feedback}

{Q}-learning is typically defined to expect the feedback to be the expected one-step reward $f_t = R(s_t,a_t)$ or a value whose expectation is $R(s_t,a_t)$. It has been proven to converge to optimal behavior under this type of feedback~\cite{Watkins92,littman96,singh00,qlearningConvergence}.  See figure \ref{graphReward}.


\subsubsection{Policy Feedback}

Given policy feedback, {Q}-learning will optimize the expected sum of future ``rewards'', which, in this case, is an indicator of whether the agent's selected action is the trainer's target policy or not.


Policy feedback depends on only the previous state and action, and, as such, {Q}-learning can treat this feedback as a reward function and converge on the behavior that optimizes the sum of these feedbacks.


Interestingly, the policy that optimizes the sum of policy feedbacks is exactly the target policy. This observation follows from the fact that matching the trainer's target policy results in a value of $1 + \gamma^1 + \gamma^2 + \gamma^3 + \cdots$. On the other hand, selecting even a single action that does not match the trainer's target policy results in the removal of one of these terms and therefore lower value. Under policy feedback, {Q}-learning thus converges to the policy that matches the trainer's target policy.  See figure \ref{graphPolicy}.




\subsubsection{Advantage Feedback}

{Q}-learning is not designed to work with advantage feedback because the advantage function is policy dependent and can cause its reward signals to change as it updates its value. Nevertheless, advantage feedback does provide a signal for how values should \emph{change} and, empirically, we often see {Q}-learning handling advantage feedback well. The analytical challenge is that the changes in the policy influence the reward and the changes in the reward influence the policy, so these two functions need to converge \emph{together} for {Q}-learning to handle advantage feedback successfully.

We conjecture that careful annealing of {Q}-learning's learning rate could provide a mechanism for stabilizing these two different adaptive processes. 
Resolving this question is a topic for future work. We believe the work done by \citeauthor{borkar} (\citeyear{borkar}) could provide greater insight.  See figure \ref{graphAdvantage}, where {Q}-learning appears to converge for a simple GridWorld domain.

\section{Conclusion}
In this paper, we analyzed the convergence of COnvergent Actor-Critic by Humans \cite{coach} under three types of feedback---one-step reward, policy, and advantage feedback. These are all examples of feedback a human trainer might give.

We defined a COACH variant called E-COACH and demonstrated its convergence under these types of feedback.  Original COACH, unfortunately, does not necessarily converge to an optimal policy under the feedback types defined in this paper.  In addition, we compared the new E-COACH with two algorithms: {Q}-learning and TAMER.
TAMER does poorly under one-step-reward feedback.
And {Q}-learning appears to converge to optimal behavior under one-step-reward and policy feedback, but future work is required to determine its performance under advantage feedback.

\bibliography{content/refs}
\bibliographystyle{icml2021}

\end{document}